# 一個基於機器學習方法的運動疲勞偵測模型

吳銘晏[1]　陳志華[2,*]　羅濟群[3]

[1] 中華電信研究院

[2] 國立高雄科技大學資訊管理系

[3] 國立交通大學資訊管理與財務金融學系

[*]通訊作者：陳志華(Email: chihua0826@gmail.com)

# 摘要

本研究提出一個基於即時生理資料的運動疲勞偵測模型，結合時域分析(time domain analysis)、頻域分析(frequency domain analysis)、去趨勢波動分析(detrended fluctuation analysis, DFA)、近似熵(approximate entropy, ApEn)與樣本熵(sample entropy, SampEn)。此外，本研究提出一個結合層次分析法(analytical hierarchy process, AHP)的特徵因子分析方法，用以萃取出關鍵影響因子。最後，再運用機器學習演算法分析各個資料特徵後辨別個人的運動疲勞程度。根據實證結果顯示，本研究提出的運動疲勞偵測模型結合特徵萃取方法後可有效辨別運動疲勞程度，並將準確率提升至98.65%。

# An Exercise Fatigue Detection Model Based on Machine Learning Methods

Ming-Yen Wu [1]　Chi-Hua Chen [2,*]　Chi-Chun Lo [3]

[1] Telecommunication Laboratories, Chunghwa Telecom Co., Ltd.

[2] Department of Information Management, National Kaohsiung University of Science and Technology

[3] Department of Information Management and Finance, National Chiao Tung University

[*]corresponding author: Chi-Hua Chen (Email: chihua0826@gmail.com)

# Abstract

This study proposes an exercise fatigue detection model based on real-time clinical data which includes time domain analysis, frequency domain analysis, detrended fluctuation analysis, approximate entropy, and sample entropy. Furthermore, this study proposed a feature extraction method which is combined with an analytical hierarchy process to analyze and extract critical features. Finally, machine learning algorithms were adopted to analyze the data of each feature for the detection of exercise fatigue. The practical experimental results showed that the proposed exercise fatigue detection model and feature extraction method could precisely detect the level of exercise fatigue, and the accuracy of exercise fatigue detection could be improved up to 98.65%.

# 壹、前言

近年來隨著慢跑活動的盛行和物聯網技術的普及，許多民眾參與慢跑運動，並穿載物聯網設備即時監測生理狀態，分析自己的運動狀況。其中，在運動過程中身體會由交感神經系統主導並分泌腎上腺素，加快心臟跳動；而運動後則是由副交感神經系統主導使肌肉鬆弛，減緩心臟跳動。隨著運動強度的不同，交感神經系統與副交感神經系統也會呈現不一樣的效果。若是發生運動過度情況，可能產生運動傷害，也容易引起交感神經與副交感神經系統的失衡，進而引發運動疲勞狀況產生(Hedelin, Wiklund, Bjerle, & Henriksson-Larsén, 2000)。因此，在運動強度和運動疲勞狀態的偵測上，可經由穿載心率監測裝置(即物聯網設備)，收集使用者的即時心率(heart rate, HR)、心跳間期(R wave-to-R wave interval, RR-interval)等資訊，再透過心率變異(heart rate variability, HRV)分析，以有效評估生理的自律神經系統(autonomic nervous system, ANS)(即交感神經系統和副交感神經系統)和估計運動狀況(Freeman, Dewey, Hadley, Myers, Froelicher, 2006)。

現有的心率監測裝置主要僅是紀錄當下的心率值與心跳間期訊號，並呈現歷史資料。而單純的心率資料僅了解在運動過程中或非運動狀態當下的心率值，而無法分析使用者運動強度和疲勞狀態。然而，根據研究指出心率在醫學上可用來評估個人生理狀況，每一次的心臟跳動都傳達了許多訊息(Vargas & Marino, 2014)。因此，需要將歷史的心率與心跳間期訊號進行 HRV 分析，運用線性或非線性分析方式，提取 HRV 量測指標，分析使用者即時生理資料且不論動態或靜態情境下，評估生理狀況與自律神經系統的變化。此外，HRV 分析相關文獻多為僅以時域分析(time domain analysis)或是頻域分析(frequency domain analysis)為基礎，兩者皆屬於線性分析方式(linear analysis)(Penzel, Kantelhardt, Grote, Peter, & Bunde, 2003)。近幾年開始有學者提出非線性分析方式(nonlinear analysis)，包含有去趨勢波動分析(detrended fluctuation analysis, DFA)(Baumert, Javorka, Seeck, Faber, Sanders, & Voss, 2012)、近似熵(approximate entropy, ApEn)(Tsuji, Suzuki, Hitomi, Yoshida, & Mizuno-Matsumoto, 2016)與樣本熵(sample entropy, SampEn)(Aktaruzzaman & Sassi, 2014)等方法(Richman & Moorman, 2000; Tarvainen, Niskanen, Lipponena, Ranta-ahoa, & Karjalainen, 2014)，可用以統計和分析心率異常時的特徵值，以及過濾雜訊，改進線性分析方式不足之處。

有鑑於此，本研究提出一個基於即時生理資料的運動疲勞偵測模型，結合時域分析、頻域分析、去趨勢波動分析、近似熵與樣本熵。此外，本研究提出一個結合層次分析法(analytical hierarchy process, AHP)的特徵因子分析方法，用以萃取出關鍵影響因子。最後，再運用機器學習演算法分析各個資料特徵後辨別個人的運動疲勞程度。

此論文以下分為五個章節，在第二節中將探討心率變異分析方法相關的技術背景。第三節說明本研究所提出之基於即時生理資料的運動疲勞偵測模型。第四節將實作本研究提出之系統與方法，並針對系統效能進行實證和分析。最後一節則說明此論文之結論與未來研究方向。

# 貳、文獻探討

心率變異分析用來評估人體的自律神經系統的活性指標。其中，在連續心電圖中的 QRS 波(Q 波、R 波、以及 S 波)之間隔裡，其相鄰的 R 波代表著心跳週期，此間隔即為 RR-interval (即心跳間期)，如圖 1 所示。

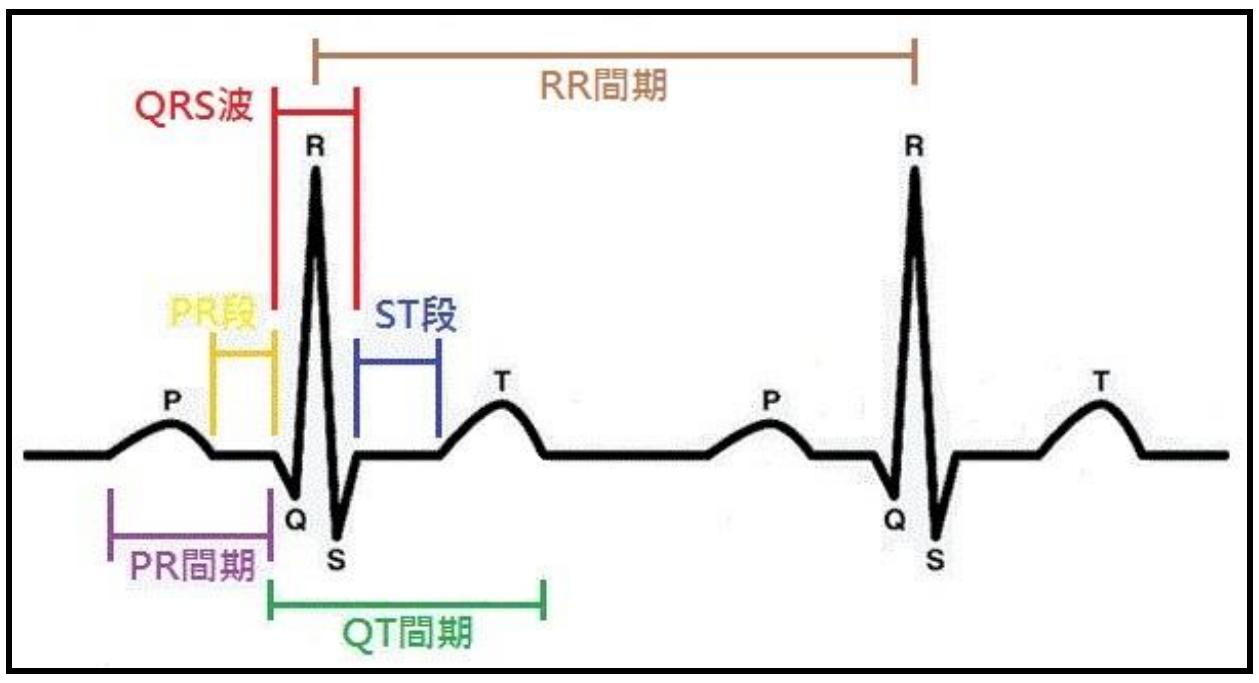

圖 1 心電圖波形

目前 HRV 以兩種分析方式進行，線性分析方式(例如：時域分析與頻域分析)和非線性分析方式(例如：去趨勢波動分析、近似熵與樣本熵)。其中，線性分析方式輸入的生理資料(如：RR-interval)，經過線性分析後，與輸出訊號之間存在線性比例的關係。而非線性分析方式則代表所輸入的生理資料經過非線性分析處理後，與輸出訊號之間不維持線性比例的關係。

## 一、線性分析方式─時域分析

時域分析主要是以統計方式計算 RR-interval 時間序

列之變異程度，而一般偵測時間是以 5 分鐘或 24 小時為基準。其中，HR ($h$ BPM (beats per minutes))與 RR-interval ($r$ 毫秒)之間的關係如公式(1)所示，共可整理為下列七種量測指標。

$$r = \frac{60}{h} \times 1000 \qquad (1)$$

(1). MeanHR (mean of all normal heart rate)：表示全部正常心率之平均值，單位為 BPM。

(2). MeanRR (mean of all normal to normal interval)：表示全部正常心跳間期之平均值，單位為毫秒。

(3). SDHR (standard deviation of all normal heart rate)：表示全部正常心率之標準差，單位為 BPM。

(4). SDNN (standard deviation of all normal to normal interval, SDNN)：表示全部正常心跳間期(normal to normal intervals, NN intervals)之標準差，單位為毫秒。

(5). R_MSSD (the square root of the mean of the sum of the squares of differences between adjacent normal to normal intervals)：表示正常心跳間期差值平方和的均方根，單位為毫秒。

(6). NN50 (the number of pairs of adjacent NN intervals differing by more than 50 ms in the entire recording)：表示所有每對相鄰正常心跳間期差距超過 50 毫秒的個數。

(7). pNN50 (NN50 count divided by the total number of all NN intervals)：表示 NN50 除以所有正常心跳間期總數，單位為百分比。

在上述時域分析之量測指標中，R-MSSD、NN50 與 pNN50 都屬於短期的變異度指標，三者之間呈現高度相關性，用來評估 HRV 中高頻的變異程度。

## 二、線性分析方式─頻域分析

頻域分析是透過傅立葉轉換(Fourier transformation, FT)將 RR-interval 時間序列轉換為頻域，以功率頻譜密度(Power spectral density, PSD)方式表現，共可整理為下列七種量測指標：

(1). 總功率(Total power, TP)：頻譜範圍位在≤0.4 Hz，為全部正常心跳間期之變異數，亦為高頻、低頻與極低頻之變異數總和，主要用來評估整體的心跳間期變異程度。

(2). 高頻功率(High frequency, HF)：頻譜範圍位在 0.15 - 0.4 Hz，為高頻範圍的正常心跳間期之變異數，代表副交感神經活性指標。

(3). 低頻功率(Low frequency, LF)：頻譜範圍位在 0.04 - 0.15 Hz，為低頻範圍的正常心跳間期之變異數，代表交感神經活性或交感神經與副交感神經同時調控的指標。

(4). 極低頻功率(Very low frequency, VLF)：頻譜範圍位在 0.003 - 0.04 Hz，為極低頻範圍的正常心跳間期之變異數。

(5). 標準化高頻功率(Normalized HF, nHF)：指 $\frac{HF}{TP - VLF} \times 100$，代表副交感神經活性的定量指標。

(6). 標準化低頻功率(Normalized LF, nLF)：指 $\frac{LF}{TP - VLF} \times 100$，代表交感神經活性的定量指標。

(7). 低、高頻功率之比值：指 $\frac{LF}{HF}$，代表交感神經與副交感神經平衡的指標。

## 三、非線性分析方式

在非線性分析方式中主要包含有：去趨勢波動分析方法、近似熵方法、以及樣本熵方法(Richman & Moorman, 2000; Tarvainen et al., 2014)，分述如下。

去趨勢波動分析方法主要將設定一固定長度值，把收集到的生理資料依該固定長度值進行切割得到數個片段。之後再計算每個片段中的資料之平均值與標準差，依此取得資料的趨勢波動(Baumert et al., 2012)。此方法的優點是可以取得整個片段資料的趨勢波動，但其限制則在於可能容易受到偏差值(outliers)的影響。

近似熵方法是由學者 Pincus 提出的一種量化序列的複雜性研究，可以用來分析一序列的資料亂度表現(Pincus, Gladstone, & Ehrenkranz, 1991)。其中，主要將設定一資料維度值和一門檻值，把收集到的生理資料依該資料維度值進行切割得到數個片段。之後再計算每個片段中的資料高於門檻值的數量和所有資料量的比例作為資料亂度。再計算不同資料維度時所得到的資訊量即為 ApEn，其值介於 0-1 之間，當 ApEn 值愈小時，其亂度愈不明顯，序列中的資料變異程度愈小，反之則變異程度愈大。此方法的主要優點是不易受到偏差值的影響，但其限制則在於門檻值的設定將依案例而異。

樣本熵方法是由學者 Richman 和 Moorman 於 2000 年時提出的近似熵方法修正模型，並與近似熵方法相似

可計算一序列的資料亂度表現(Richman & Moorman, 2000)。經由資料維度值和門檻值等設定後，再把生理資料依該資料維度值進行切割得到數個片段。再運用門檻值計算每個片段的資料亂度。而樣本熵方法主要考量不同資料維度時資料亂度的比例，故更能表現出增加資料維度時資料亂度降低的倍數。此方法的主要優點為取得資料亂度比例，並可不受偏差值影響；但最大限制在於當資料亂度為 0 時，此方法將不適用。

有鑑於線性分析方式(例如：時域分析與頻域分析)和非線性分析方式(例如：去趨勢波動分析、近似熵與樣本熵)分別各有其適用的資料分佈，故在本研究中將分別採用上述線性分析方式和非線性分析方式計算生理資料，並分別產生不同資料特徵作為運動疲勞偵測模型使用。

# 參、基於即時生理資料的運動疲勞偵測模型

為了能提高辨別運動疲勞狀態的準確度，本節提出了一個基於即時生理資料的運動疲勞偵測模型，並且提出一個特徵萃取方法結合層次分析法，期望從五種 HRV 量測指標中萃取出辨識能力較佳的因子。在 3.1 節會先定義欲解決的問題與模型，再於 3.2 節詳述演算法設計步驟。最後再於 3.3 節討論本研究方法的特色和限制。

## 一、問題定義

在 HRV 分析上，如同第二節所提及，可分為時域分析(包含 7 種量測指標)、頻域分析(包含 7 種量測指標)、去趨勢波動分析方法(包含 2 種量測指標，分別為高頻功率標準差和低頻功率標準差)、近似熵方法(包含 1 種量測指標)、以及樣本熵方法(包含 1 種量測指標)，共 18 種量測指標。在本研究中，雖以五種 HRV 分析方式進行運動疲勞偵測，但仍會遇到幾項需要探討的議題：

(1). 如何從 18 種 HRV 量測指標中萃取出具有高辨識能力的量測指標，以改善辨識準確度。

(2). 經過本研究方法所萃取後的 HRV 量測指標，各個量測指標間的相關性探討。

(3). 如何以運動時間與生理資料衡量當下或未來運動疲勞狀態。

因此，本研究提出一套新穎的特徵萃取方法，經由分析兩兩因子的資料分佈來判斷其相依性，再搭配層次分析法萃取出重要特徵因子，以建立運動疲勞偵測模型。本研究方法預期將改善原有文獻中僅以單一的時域分析方法，幫助使用者辨別不同運動狀態時的運動疲勞狀態，並降低誤判的機率，提高辨識準確度。

本研究所使用的相關參數定義如表 1 所示。

表 1 參數名詞定義

| 符號 | 定義 |
|---|---|
| $\overline{HR}$ | 全部正常心率之平均值。 |
| $\overline{RR}$ | 全部正常心跳間期之平均值。 |
| *SDHR* | 全部正常心率之標準差。 |
| *SDNN* | 全部正常心跳間期之標準差。 |
| *R_MSSD* | 正常心跳間期差值平方和的均方根。 |
| *NN50* | 所有每對相鄰正常心跳間期差距超過 50 毫秒的個數。 |
| *pNN50* | NN50 除以所有正常心跳間期總數。 |
| $\sigma$ | 常態分佈 $\phi(x\mid\mu,\sigma^2)$ 的標準差。 |
| $\mu$ | 常態分佈 $\phi(x\mid\mu,\sigma^2)$ 的平均值。 |
| $\Phi(x,p\mid\mu,\sigma^2)$ | 以平均值 $\mu$ 與標準差 $\sigma$ 的常態分佈之累積分佈函數。 |
| $p_1$ 與 $p_2$ | 常態分佈 $\phi(x\mid\mu_1,\sigma_1^2)$ 與 $\phi(x\mid\mu_2,\sigma_2^2)$ 的相交點。 |
| *A* | 常態分佈 $\phi(x\mid\mu_1,\sigma_1^2)$ 與 $\phi(x\mid\mu_2,\sigma_2^2)$ 的交集面積。 |
| $a_1,a_2,\ldots,a_n$ | 一集合 *n* 的比較屬性。 |
| $w_1,w_2,\ldots,w_n$ | 比較屬性 *a* 的權重。 |
| $g_i$ | 第 *i* 類別的規模大小。 |
| $r_{n,G}$ | 規模大小 *g* 的兩兩相加組合。 |
| $M_G$ | 規模大小 *g* 的兩兩相加組合之矩陣。 |
| $\omega_k$ | 第 *k* 個特徵的重要價值。 |
| $W_F$ | 權重矩陣。 |

## 二、研究方法設計與原理

本節將先介紹整體服務架構，再說明運動疲勞偵測模型之建置，最後再說明實施流程，分述如下。

### (一)、服務架構

本研究之服務架構主要包含：(1) 胸戴式心率帶、(2) 智慧型行動裝置、(3) HRV 生醫訊號分析平台、(4) 運動疲勞偵測模型，如圖 2 所示。

(1). 胸戴式心率帶：透過片導電橡膠偵測心率，心臟跳動時所產生的微量電位差經過放大處理後所得到的訊號，再透過低功耗藍牙傳輸至接收裝置。

(2). 智慧型行動裝置：具有低功耗藍牙 4.0 以上之裝置接收心率帶所傳輸的生醫訊號，如 HR 與 RR-interval。經過每 5 分鐘的訊號收集，再由無線網路傳輸至分析平台進行 HRV 分析。

(3). HRV 生醫訊號分析平台：以五種 HRV 分析進行訊號分析，並將分析完的 HRV 量測指標與運動疲勞偵測模型進行狀態辨識。

(4). 運動疲勞偵測模型：用以辨別 HRV 生醫訊號所對應的運動疲勞狀態，於 3.2.2 節中詳細介紹。

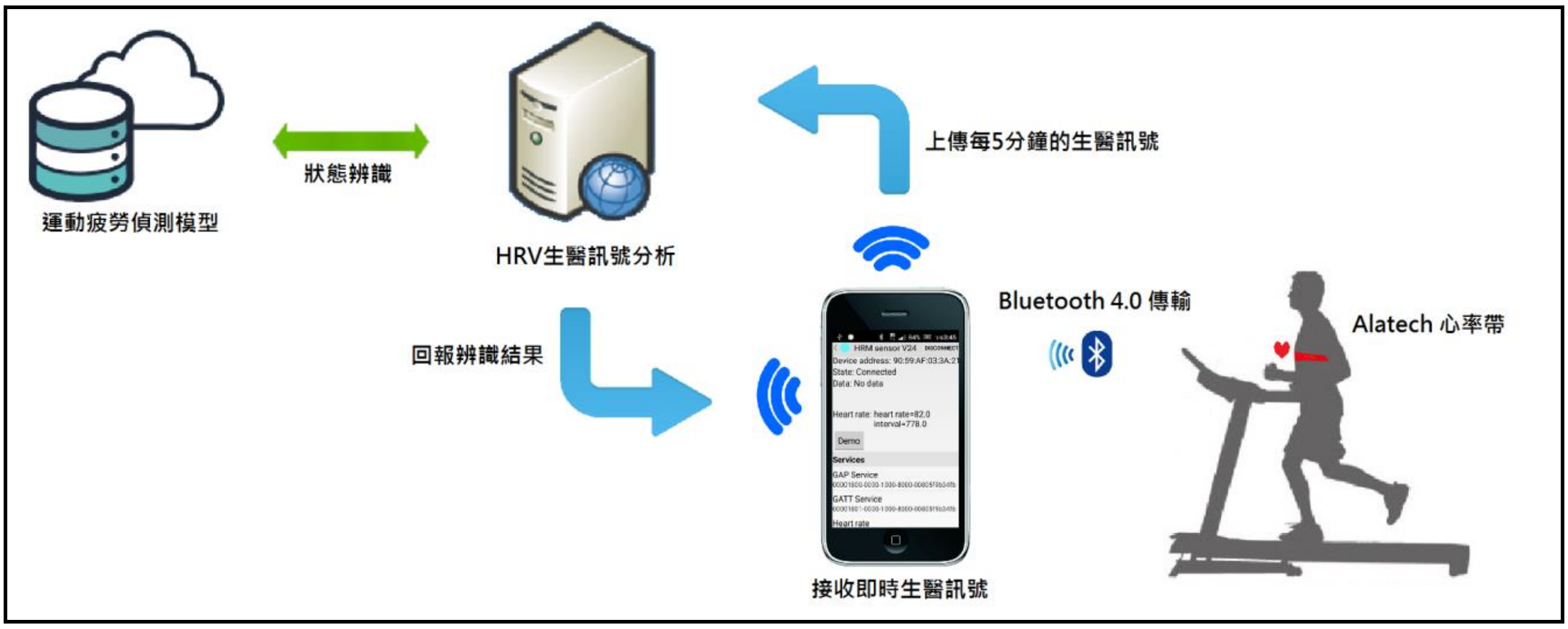

圖 2 服務架構圖

**(二)、運動疲勞偵測模型之建置**

本研究設計一運動疲偵測模型，其建置流程如圖 3 所示。主要分為六個步驟進行，依序是：(1) 生醫訊號偵測前準備、(2) 進行五種不同運動狀態與生醫訊號偵測、(3) 進行五種不同 HRV 訊號分析、(4) 量測兩個常態分佈特徵因子的交集面積、(5) 基於層次分析法之 HRV 量測指標權重計算、(6) 進行機器學習分類模型訓練與分類模型建置，如下述說明。

**1. 生醫訊號偵測前準備**

本研究實際偵測在不同運動狀態下的生醫訊號，而在偵測生醫訊號前準備如下敘述：

(1). 運動前：在開始運動前的 60 分鐘即開始量測生醫訊號，以每 5 分鐘為一筆量測單位。

(2). 緩和與劇烈運動中：在開始運動後的前 3 分鐘停止測量，3 分鐘後開始量測生醫訊號，以每 5 分鐘為一筆量測單位。

(3). 緩和與劇烈運動後：在停止運動後的前 3 分鐘停止測量，3 分鐘後開始量測生醫訊號，以每 5 分鐘為一筆量測單位。

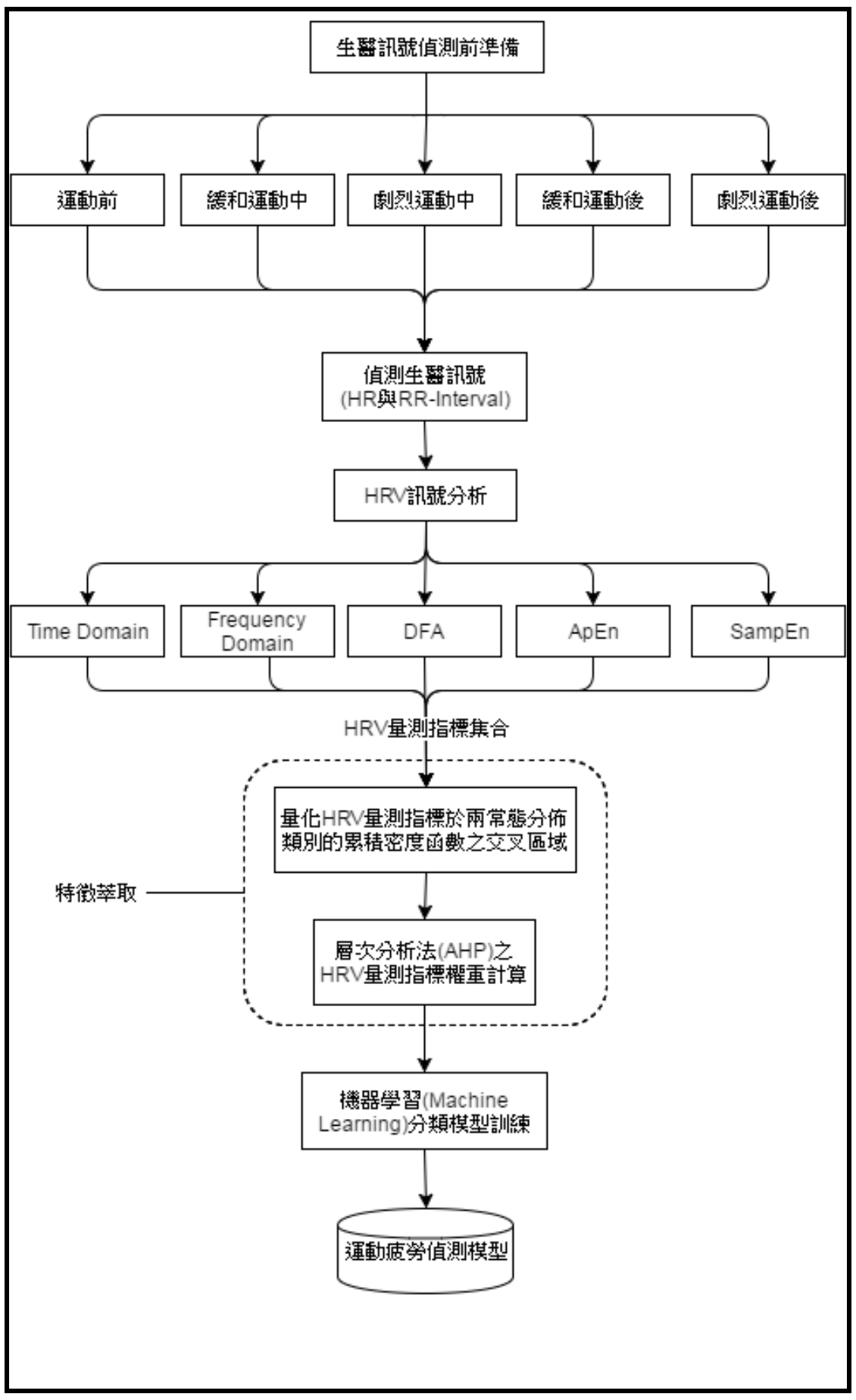

圖 3 運動疲勞偵測模型建置流程圖

**2. 進行五種不同運動狀態與生醫訊號偵測**

根據美國疾病與預防中心(Centers for disease control and prevention, CDC)定義一個人的最大心率值為：220-現在年齡。而 CDC 亦定義了不同運動狀態下的心率表現，緩和(中度)運動和劇烈(重度)運動之之心率表現分別如公式(2)和(3)所示。以一位 20 歲年輕人為例，其緩和運動的心率值界於 100/bmp-140/bpm 之間，而劇烈運動的心率值則大於 140/bmp。

$$(220 - \text{現在年齡}) \times 50\% \le HR \le (220 - \text{現在年齡}) \times 70\% \quad (2)$$

$$HR > (220 - \text{現在年齡}) \times 70\% \quad (3)$$

本研究在開始運動前的行為是以靜坐方式量測；緩和運動的行為是以慢跑、快走與慢速踩踏飛輪方式量測；劇烈運動的行為則是在快跑與快速踩踏飛輪方式量測；緩和與劇烈運動後的行為是以站立或靜坐方式量測。

**3. 進行五種不同 HRV 訊號分析**

本研究 HRV 分析包含時域分析、頻域分析、去趨勢波動分析(DFA)、近似熵與樣本熵。以下本研究將以數學式表示 2.1 節所提到之時域分析中七種量測指標的計算方式，如公式(4)~(10)所示，其中 $n$ 為總筆數。

$$MeanHR = \overline{HR} = \frac{1}{n}\sum_{i=1}^{n} HR_i \quad (4)$$

$$MeanRR = \overline{RR} = \frac{1}{n}\sum_{i=1}^{n} RR_i \quad (5)$$

$$SDHR = \sqrt{\frac{\sum_{i=1}^{n}\left(HR_i - \overline{HR}\right)^2}{n}} \quad (6)$$

$$SDNN = \sqrt{\frac{\sum_{i=1}^{n}\left(RR_i - \overline{RR}\right)^2}{n}} \quad (7)$$

$$R_MSSD = \sqrt{\frac{\sum_{i=1}^{n-1}\left(RR_{i+1} - RR_i\right)^2}{n-1}} \quad (8)$$

$$NN50 = \sum_{i=1}^{n} c_i, c_i = \begin{cases} 1, if \ (RR_{i+1} - RR_i) \ge 50 \\ 0, \text{otherwise} \end{cases} \quad (9)$$

$$pNN50 - \frac{NN50}{n} \quad (10)$$

另外，頻域分析透過傅利葉轉換為功率頻譜密度，並取得不同頻率底下的功率值，分別有 TP、HF、LF、VLF、nHF、nLF 與 LF/HF 一共七個量測指標。而非線性分析方式有去趨勢波動分析方法(包含 2 種量測指標，分別為高頻功率標準差和低頻功率標準差)、近似熵方法(包含 1 種量測指標)。

**4. 量測兩個常態分佈特徵因子的交集面積**

本節假設特徵因子的資料分佈呈現常態分佈，並提出計算常態分佈交集面積的計算方式，當交集面積越小則代表兩因子之間的重疊越少，越有助於分類使用，藉此挑選出重要的特徵因子。首先，將證明各個 HRV 量測指標皆為常態分佈，之後再介紹如何計算兩個常態分佈的交集面積。

**(1). 各個 HRV 量測指標皆為常態分佈之證明**

在提取 18 種 HRV 量測指標後，每個量測指標所對應到的五種不同運動狀態類別也會呈現不同的資料分佈狀況。因此，在量測兩個常態分佈特徵因子的交集面積前，本研究必須先檢驗每種類別的因子分佈是否皆為常態分佈。

根據卡方分配表得到在顯著水準 $\alpha = 0.05$，自由度 $df = 10$，卡方臨界值 $\chi^2 = 18.307$，而在運動前之心率資料分佈與常態分佈的檢定在顯著水準 $\alpha = 0.05$，自由度 $df = 10$，其卡方統計量值為 0.982，小於卡方臨界值。因此，運動前之心率分佈符合常態分佈。表 2 為各個類別心率資料分佈的卡方檢定。根據結果顯示，每個類別的資料分佈皆符合常態分佈。因此，本研究可以根據所有 HRV 量測指標在各個類別的常態分佈進行交集面積計算。

表 2 各個類別心率資料分佈的卡方檢定

| 類別 | 卡方檢定統計量值 | 自由度 $df$ | 顯著水準 $\alpha$ | 卡方臨界值 $\chi^2$ | 符合常態分佈 |
|---|---|---|---|---|---|
| 運動前 | 0.982 | 10 | 0.05 | 18.307 | 符合 |
| 劇烈運動中 | 0.72 | 15 | 0.05 | 24.996 | 符合 |
| 緩和運動中 | 16.64 | 11 | 0.05 | 19.675 | 符合 |
| 劇烈運動後 | 0.59 | 10 | 0.05 | 18.307 | 符合 |
| 緩和運動後 | 1.81 | 3 | 0.05 | 7.815 | 符合 |

當所有 HRV 量測指標在各個類別的常態分佈形成

一交集面積時，若交集面積愈大，則此量測指標影響辨識結果的程度愈小；反之交集面積愈小，則此量測指標影響辨識結果的程度愈大。如圖 4(a)與圖 4(b)所示，本研究發現特徵 A 在類別 1 與類別 2 的常態分佈其交集部分較小，兩常態分佈重疊部分較少；而特徵 B 在類別 1 與類別 2 的常態分佈其交集部分較大，兩常態分佈重疊部分較多。因此本研究可以說特徵 A 較特徵 B 在類別 1 與類別 2 的辨識程度較好。

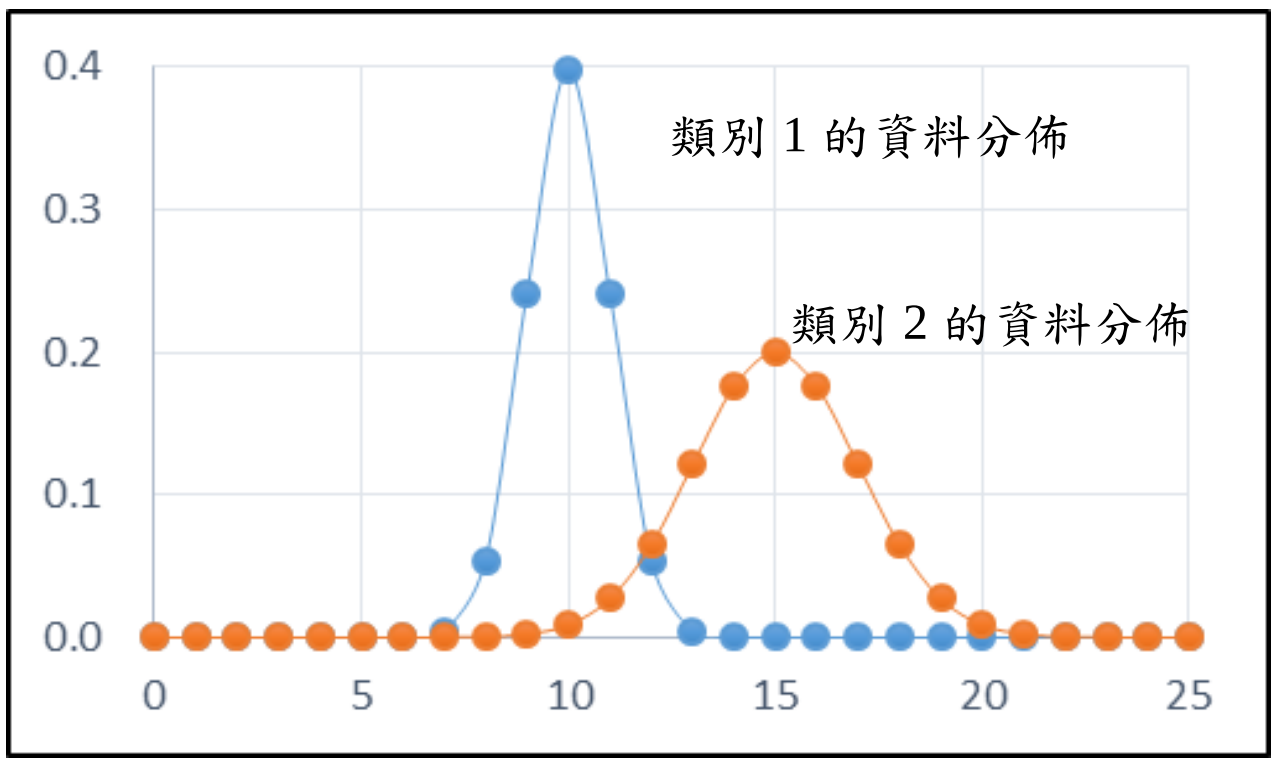

圖 4(a) 特徵 A 在類別 1、2 的常態分佈狀況

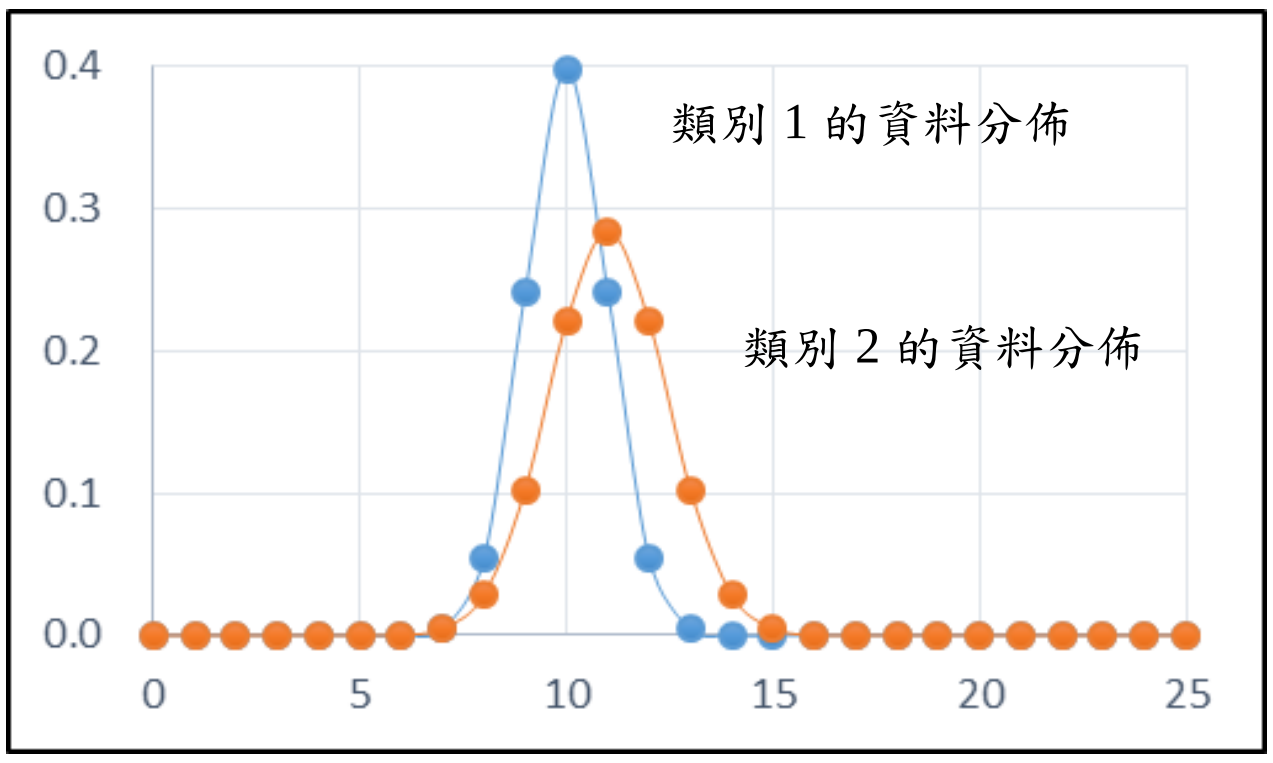

圖 4(b) 特徵 B 在類別 1、2 的常態分佈狀況

**(2). 兩個常態分佈交集面積之計算**

假設類別 1 的資料為常態分佈，其函數為 $\phi(x \mid \mu_1, \sigma_1^2)$、標準差為 $\sigma_1$、平均值為 $\mu_1$，如公式(11)所示。類別 2 的資料為常態分佈，其函數為 $\phi(x \mid \mu_2, \sigma_2^2)$、標準差為 $\sigma_2$、平均值為 $\mu_2$，如公式(12)所示。因此，可計算兩個常態分佈的交點，令 $\phi(x \mid \mu_1, \sigma_1^2) = \phi(x \mid \mu_2, \sigma_2^2)$，如公式(13)。

當 $x$ 值求出後，令這兩個交點定義為 $p_1$ 與 $p_2$，如公式(14)與(15)所示。又常態分佈的累積分佈函數和誤差函數(error function)，分別定義如公式(16)和(17)所示。類別 1 和類別 2 資料分佈的交集面積可經由公式(18)計算得到。

$$\phi(x \mid \mu_1, \sigma_1^2) = \frac{1}{\sigma_1\sqrt{2\pi}} e^{-\frac{(x-\mu_1)^2}{2\sigma_1^2}} \tag{11}$$

$$\phi(x \mid \mu_2, \sigma_2^2) = \frac{1}{\sigma_2\sqrt{2\pi}} e^{-\frac{(x-\mu_2)^2}{2\sigma_2^2}} \tag{12}$$

$$\begin{aligned} &\phi(x \mid \mu_1, \sigma_1^2) = \phi(x \mid \mu_2, \sigma_2^2) \\ &\Rightarrow \frac{1}{\sigma_1\sqrt{2\pi}} e^{-\frac{(x-\mu_1)^2}{2\sigma_1^2}} = \frac{1}{\sigma_2\sqrt{2\pi}} e^{-\frac{(x-\mu_2)^2}{2\sigma_2^2}} \\ &\Rightarrow x = \frac{\mu_2\sigma_1^2 - \mu_1\sigma_2^2 \pm \sqrt{2\sigma_1^4\sigma_2^2 \ln\left(\frac{\sigma_1}{\sigma_2}\right) + \sigma_1^2\mu_2^2\sigma_2^2 + \mu_1^2\sigma_1^2\sigma_2^2 - 2\sigma_1^2\sigma_2^4 \ln\left(\frac{\sigma_1}{\sigma_2}\right) - 2\mu_1\mu_2\sigma_1^2\sigma_2^2}}{(\sigma_1^2 - \sigma_2^2)} \end{aligned} \tag{13}$$

$$p_1 = \frac{\mu_2\sigma_1^2 - \mu_1\sigma_2^2 - \sqrt{2\sigma_1^4\sigma_2^2 \ln\left(\frac{\sigma_1}{\sigma_2}\right) + \sigma_1^2\mu_2^2\sigma_2^2 + \mu_1^2\sigma_1^2\sigma_2^2 - 2\sigma_1^2\sigma_2^4 \ln\left(\frac{\sigma_1}{\sigma_2}\right) - 2\mu_1\mu_2\sigma_1^2\sigma_2^2}}{(\sigma_1^2 - \sigma_2^2)} \tag{14}$$

$$p_2 = \frac{\mu_2\sigma_1^2 - \mu_1\sigma_2^2 + \sqrt{2\sigma_1^4\sigma_2^2 \ln\left(\frac{\sigma_1}{\sigma_2}\right) + \sigma_1^2\mu_2^2\sigma_2^2 + \mu_1^2\sigma_1^2\sigma_2^2 - 2\sigma_1^2\sigma_2^4 \ln\left(\frac{\sigma_1}{\sigma_2}\right) - 2\mu_1\mu_2\sigma_1^2\sigma_2^2}}{(\sigma_1^2 - \sigma_2^2)} \tag{15}$$

$$\Phi\left(x,p\mid\mu,\sigma^2\right)=\int_{x=-\infty}^{p}\frac{1}{\sigma\sqrt{2\pi}}e^{-\frac{(x-\mu)^2}{2\sigma^2}}=\frac{1}{2}\left[1+\operatorname{erf}\left(\frac{p-\mu}{\sigma\sqrt{2}}\right)\right] \tag{16}$$

$$\begin{aligned}
&\operatorname{erf}(z)\approx 1-t(z)\\
&\quad\text{where } t(z)=\frac{1}{\left(\sum_{i=0}^{6}a_i z^i\right)^{16}}, a_0=1, a_1=0.0705230784, a_2=0.0422820123,\\
&\qquad a_3=0.0092705272, a_4=0.0001520143, a_5=0.0002765672, a_6=0.0000430638
\end{aligned} \tag{17}$$

$$\begin{aligned}
A&=\left[\Phi\left(x,p_2\mid\mu_1,\sigma_1^2\right)\right]+\left[\Phi\left(x,p_1\mid\mu_2,\sigma_2^2\right)-\Phi\left(x,p_2\mid\mu_2,\sigma_2^2\right)\right]+\left[1-\Phi\left(x,p_1\mid\mu_1,\sigma_1^2\right)\right],\ \text{where } \sigma_1<\sigma_2\\
&=\left[\Phi\left(x,p_1\mid\mu_2,\sigma_2^2\right)\right]+\left[\Phi\left(x,p_2\mid\mu_1,\sigma_1^2\right)-\Phi\left(x,p_1\mid\mu_1,\sigma_1^2\right)\right]+\left[1-\Phi\left(x,p_2\mid\mu_2,\sigma_2^2\right)\right],\ \text{where } \sigma_1>\sigma_2\\
&=1+\frac{1}{2}\left[\operatorname{erf}\left(\frac{p_2-\mu_1}{\sigma_1\sqrt{2}}\right)+\operatorname{erf}\left(\frac{p_1-\mu_2}{\sigma_2\sqrt{2}}\right)-\operatorname{erf}\left(\frac{p_2-\mu_2}{\sigma_2\sqrt{2}}\right)-\operatorname{erf}\left(\frac{p_1-\mu_1}{\sigma_1\sqrt{2}}\right)\right]\\
&\approx 1+\frac{1}{2}\left[1-t\left(\frac{p_2-\mu_1}{\sigma_1\sqrt{2}}\right)+1-t\left(\frac{p_1-\mu_2}{\sigma_2\sqrt{2}}\right)-1+t\left(\frac{p_2-\mu_2}{\sigma_2\sqrt{2}}\right)-1+t\left(\frac{p_1-\mu_1}{\sigma_1\sqrt{2}}\right)\right]\\
&\approx 1+\frac{1}{2}\left[t\left(\frac{p_1-\mu_1}{\sigma_1\sqrt{2}}\right)+t\left(\frac{p_2-\mu_2}{\sigma_2\sqrt{2}}\right)-t\left(\frac{p_1-\mu_2}{\sigma_2\sqrt{2}}\right)-t\left(\frac{p_2-\mu_1}{\sigma_1\sqrt{2}}\right)\right]
\end{aligned} \tag{18}$$

## 5. 基於層次分析法之 HRV 量測指標權重計算

當取得兩個常態分佈特徵因子的交集面積後，可將此交集面積作為層次分析法的輸入值，並根據層次分析法用以加權每個 HRV 量測指標的重要程度。例如，有一集合 $n$ 的比較屬性表示為 $a_1, a_2, \ldots, a_n$ ，且將權重定義為 $w_1, w_2, \ldots, w_n$ 。如公式(19)所示，$w_i$ 被定義成 $\sum_{v=1}^{n} a_{i,v}\times\frac{1}{\sum_{u=1}^{n}a_{u,v}}$ 。

在這個案例中，資料結果可以被分為 $n$ 個類別。因此，對於每兩個類別的 $q$ (i.e., $q=C_2^n=\frac{n(n-1)}{2}$) 組可以推算出 $m$ 個特徵。針對第 $j$ 個特徵來說，第 $k$ 組的交集面積被定義成 $A_{k,j}$ ，而 $A_{k,j}$ 以由公式(18)推算出來。另外，將第 $i$ 個類別的規模大小定義成 $G_i$ ，並且每個類別(即矩陣 $M_G$)的權重可以由公式(20)推算出來。依此類推，第 $k$ 組的每個特徵權重可由公式(21)計算得到。

對於最後的決策進行，最終的權重矩陣 $W_F$ 可以基於各組的規模大小之權重與 $m$ 特徵由公式(22)計算。然後，第 $k$ 個特徵的重要價值可以被評估為 $\omega_k$，最後取出最高的價值的特徵作為辨識疲勞狀況的因子。

$$M=\begin{bmatrix} a_{1,1} & \cdots & a_{1,j} & \cdots & a_{1,n}\\ \vdots & & \vdots & & \vdots\\ a_{i,1} & \cdots & a_{i,j} & \cdots & a_{i,n}\\ \vdots & & \vdots & & \vdots\\ a_{n,1} & \cdots & a_{n,j} & \cdots & a_{n,n}\end{bmatrix}\cong\begin{bmatrix} \frac{w_1}{w_1} & \cdots & \frac{w_1}{w_j} & \cdots & \frac{w_1}{w_n}\\ \vdots & & \vdots & & \vdots\\ \frac{w_i}{w_1} & \cdots & \frac{w_i}{w_j} & \cdots & \frac{w_i}{w_n}\\ \vdots & & \vdots & & \vdots\\ \frac{w_n}{w_1} & \cdots & \frac{w_n}{w_j} & \cdots & \frac{w_n}{w_n}\end{bmatrix}=W \tag{19}$$

$$\text{where } a_{i,j}>0, a_{i,j}=\frac{1}{a_{j,i}}, \text{and } w_i=\sum_{v=1}^{n}a_{i,v}\times\frac{1}{\sum_{u=1}^{n}a_{u,v}}$$

$$
M_G = \begin{bmatrix} \frac{(g_1+g_2)}{(g_1+g_2)} & \cdots & \frac{(g_1+g_2)}{(g_i+g_j)} & \cdots & \frac{(g_1+g_2)}{(g_{n-1}+g_n)} \\ \vdots & & \vdots & & \vdots \\ \frac{(g_i+g_j)}{(g_1+g_2)} & \cdots & \frac{(g_i+g_j)}{(g_i+g_j)} & \cdots & \frac{(g_i+g_j)}{(g_{n-1}+g_n)} \\ \vdots & & \vdots & & \vdots \\ \frac{(g_{n-1}+g_n)}{(g_1+g_2)} & \cdots & \frac{(g_{n-1}+g_n)}{(g_i+g_j)} & \cdots & \frac{(g_{n-1}+g_n)}{(g_{n-1}+g_n)} \end{bmatrix} \cong \begin{bmatrix} \frac{w_{1,G}}{w_{1,G}} & \cdots & \frac{w_{1,G}}{w_{j,G}} & \cdots & \frac{w_{1,G}}{w_{q,g}} \\ \vdots & & \vdots & & \vdots \\ \frac{w_{i,G}}{w_{1,G}} & \cdots & \frac{w_{i,G}}{w_{j,G}} & \cdots & \frac{w_{i,G}}{w_{q,G}} \\ \vdots & & \vdots & & \vdots \\ \frac{w_{q,G}}{w_{1,G}} & \cdots & \frac{w_{q,G}}{w_{j,G}} & \cdots & \frac{w_{q,G}}{w_{q,G}} \end{bmatrix} = W_g \tag{20}
$$

where $r_{1,G} = (g_1+g_2), r_{2,G} = (g_1+g_3), \ldots, r_{q,G} = (g_{n-1}+g_n)$,

$a_{i,j,G} = \frac{r_{i,G}}{r_{j,G}} > 0, a_{i,j,G} = \frac{1}{a_{j,i,G}}$, and $w_{i,G} = \sum_{v=1}^{q} a_{i,v,G} \times \frac{1}{\sum_{u=1}^{q} a_{u,v,G}}$

$$
M_k = \begin{bmatrix} \frac{(1-A_{k,1})}{(1-A_{k,1})} & \cdots & \frac{(1-A_{k,1})}{(1-A_{k,i})} & \cdots & \frac{(1-A_{k,1})}{(1-A_{k,m})} \\ \vdots & & \vdots & & \vdots \\ \frac{(1-A_{k,i})}{(1-A_{k,1})} & \cdots & \frac{(1-A_{k,i})}{(1-A_{k,i})} & \cdots & \frac{(1-A_{k,i})}{(1-A_{k,m})} \\ \vdots & & \vdots & & \vdots \\ \frac{(1-A_{k,m})}{(1-A_{k,1})} & \cdots & \frac{(1-A_{k,m})}{(1-A_{k,i})} & \cdots & \frac{(1-A_{k,m})}{(1-A_{k,m})} \end{bmatrix}
$$

$$
= \begin{bmatrix} \frac{r_{1,k}}{r_{1,k}} & \cdots & \frac{r_{1,k}}{r_{i,k}} & \cdots & \frac{r_{1,k}}{r_{m,k}} \\ \vdots & & \vdots & & \vdots \\ \frac{r_{i,k}}{r_{1,k}} & \cdots & \frac{r_{i,k}}{r_{i,k}} & \cdots & \frac{r_{i,k}}{r_{m,k}} \\ \vdots & & \vdots & & \vdots \\ \frac{r_{m,k}}{r_{1,k}} & \cdots & \frac{r_{m,k}}{r_{i,k}} & \cdots & \frac{r_{m,k}}{r_{m,k}} \end{bmatrix} \cong \begin{bmatrix} \frac{w_{1,k}}{w_{1,k}} & \cdots & \frac{w_{1,k}}{w_{i,k}} & \cdots & \frac{w_{1,k}}{w_{m,k}} \\ \vdots & & \vdots & & \vdots \\ \frac{w_{i,k}}{w_{1,k}} & \cdots & \frac{w_{i,k}}{w_{i,k}} & \cdots & \frac{w_{i,k}}{w_{m,k}} \\ \vdots & & \vdots & & \vdots \\ \frac{w_{m,k}}{w_{1,k}} & \cdots & \frac{w_{m,k}}{w_{i,k}} & \cdots & \frac{w_{m,k}}{w_{m,k}} \end{bmatrix} = W_k \tag{21}
$$

where $r_{1,k} = (1-A_{k,1}), r_{2,k} = (1-A_{k,2}), \ldots, r_{m,k} = (1-A_{k,m})$,

$a_{i,j,k} = \frac{r_{i,k}}{r_{j,k}} > 0, a_{i,j,k} = \frac{1}{a_{j,i,k}}$, and $w_{i,k} = \sum_{v=1}^{m} a_{i,v,k} \times \frac{1}{\sum_{u=1}^{m} a_{u,v,k}}$

$$
W_F = \begin{bmatrix} \sum_{v=1}^{q} \frac{w_{1,G}}{w_{v,G}} & \cdots & \sum_{v=1}^{q} \frac{w_{i,G}}{w_{v,G}} & \cdots & \sum_{v=1}^{q} \frac{w_{q,G}}{w_{v,G}} \end{bmatrix} \begin{bmatrix} \sum_{v=1}^{m} \frac{w_{1,1}}{w_{v,1}} & \cdots & \sum_{v=1}^{m} \frac{w_{i,1}}{w_{v,1}} & \cdots & \sum_{v=1}^{m} \frac{w_{m,1}}{w_{v,1}} \\ \vdots & & \vdots & & \vdots \\ \sum_{v=1}^{m} \frac{w_{1,i}}{w_{v,i}} & \cdots & \sum_{v=1}^{m} \frac{w_{i,i}}{w_{v,i}} & \cdots & \sum_{v=1}^{m} \frac{w_{m,i}}{w_{v,i}} \\ \vdots & & \vdots & & \vdots \\ \sum_{v=1}^{m} \frac{w_{1,q}}{w_{v,q}} & \cdots & \sum_{v=1}^{m} \frac{w_{i,q}}{w_{v,q}} & \cdots & \sum_{v=1}^{m} \frac{w_{m,q}}{w_{v,q}} \end{bmatrix} \tag{22}
$$

$$
= [\omega_1 \quad \cdots \quad \omega_k \quad \cdots \quad \omega_m] \text{ where } \omega_k = \sum_{u=1}^{q} \left( \sum_{v=1}^{q} \frac{w_{u,G}}{w_{v,G}} \times \sum_{v=1}^{m} \frac{w_{k,u}}{w_{v,u}} \right)
$$

**6. 進行機器學習分類模型訓練與分類模型建置**

本研究採用的機器學習分類演算法包含 k 個最近鄰居法(k nearest neighbors, kNN)、支持向量機(support vector machine, SVM)、貝氏分類器(naive Bayes classifier, NBC)、類神經網路(neural network, NN)與決策樹(decision tree, DT)，並針對這些方法進行比較與分析。在本研究的分類模型建置當中，依照上述層次分析法加權常態分佈交集面積，將萃取出的 HRV 量測指標作為輸入特徵，而五種不同運動狀態與疲勞狀態紀錄則為輸出結果。其中，由於資料呈現常態分佈，故對於某些監督式學習分類器(如類神經網路)有較好的分類表現。因此，在建置分類模型的過程中，較適用於辨識運動疲勞的狀態。

**(三)、運動疲勞偵測模型之流程**

本研究的運動疲勞偵測流程主要分為五個步驟進行，依序是：(1) 生醫訊號偵測前準備、(2) 生醫訊號偵測、(3) HRV 訊號分析、(4) 進行五種不同 HRV 分析、(5)依據運動疲勞偵測模型進行狀態辨識，如圖 5 所示。在步驟(1)~(4)如 3.2.2 節所述，而步驟(5)~(6)依據建立好的運動疲勞偵測模型進行狀態辨識。根據使用者當下的生醫訊號透過 HRV 分析後，與偵測模型進行分析比對，並辨別當下的 HRV 分析後的量測指標資訊是否符合運動疲勞條件，若符合運動疲勞條件，則判定為運動疲勞狀態；若不符和運動疲勞條件，則判定為非運動疲勞狀態。

## 三、方法特色和限制

本研究所提出的運動疲勞偵測模型，是依據常態分佈底下的交集面積來衡量兩常態分佈的重疊程度，再運用層次分析法進行交集面積的加權，以辨別各個 HRV 量測指標的重要程度。此方法非一般的主成份分析、多變量分析或多目標決策等方式來衡量每個因子的重要性，因此在傳統的研究中也提出了較新穎的重要特徵辨識方法。而在計算上則先找出兩常態分佈相交的兩個點，再依據交點的左右兩邊進行積分運算，以計算交集面積，因此在表達上較容易呈現，計算上也較容易執行。

有鑑於此方法必須在兩常態分佈底下計算，因此若各個特徵在兩兩類別組合的分佈出現非常態分佈，就無法使用此方式辨別各個特徵的交集面積，後續亦無法使用層次分析法進行交集面積加權，以評估量測指標的重要性。因此當資料分佈屬於指數分佈或泊松分佈等，就得設計新的交集面積計算方式。

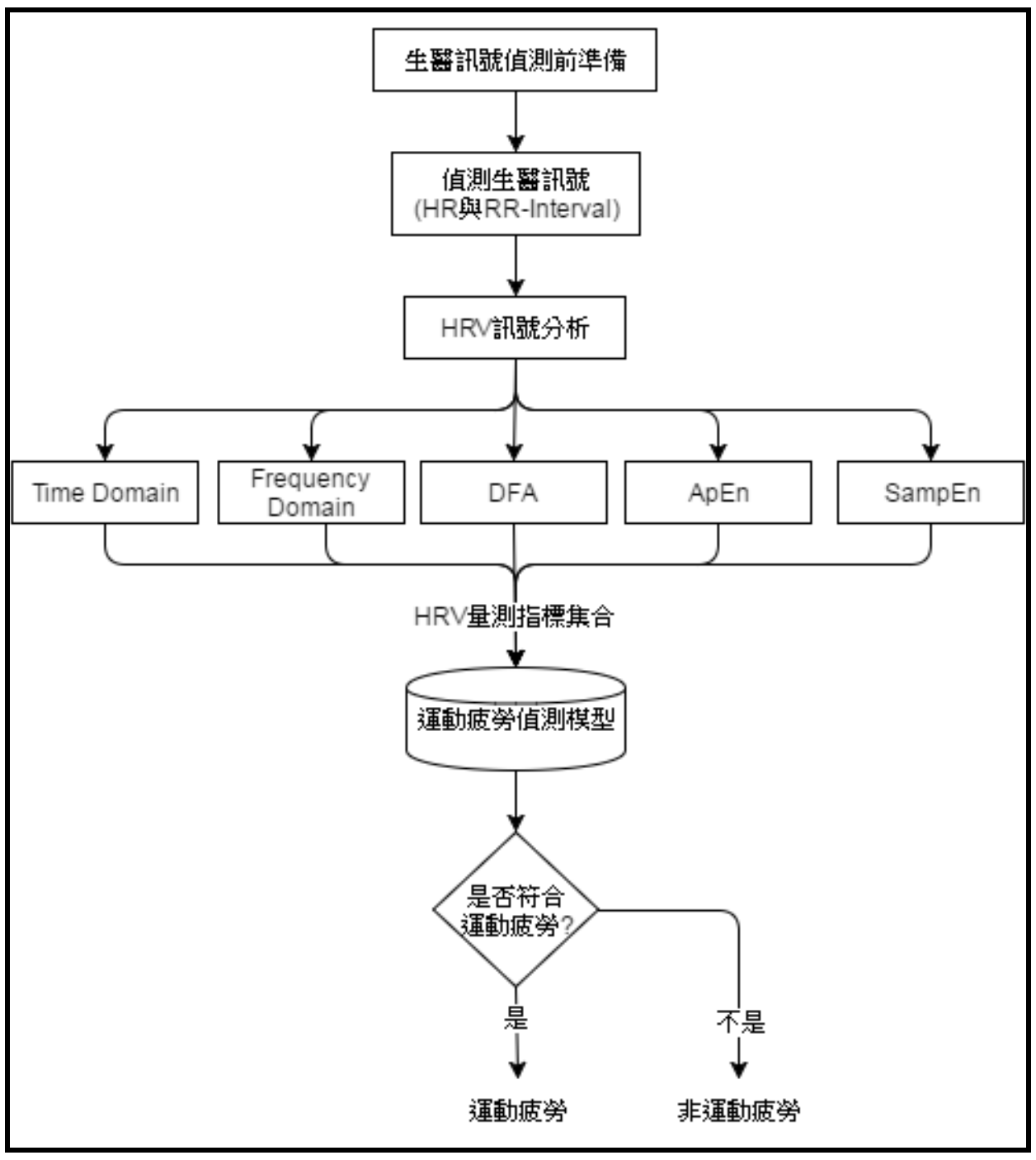

圖 5 運動疲勞偵測模型建置流程圖

# 肆、實驗與結果分析

本節將先介紹系統實作環境與資料收集數量，再針對實驗結果進行分析與討論，分述如下。

## 一、實驗環境

本節針分別對實驗平台、實驗案例與資料收集進行說明與介紹。

**(一)、實驗平台**

本研究以 Android 4.4 (API 19)建置生醫訊號偵測系統，建立與 ALATECH © - CS010 心率帶的藍牙 4.0 傳輸介面。HRV 分析平台與運動疲勞偵測模型(包含機器學習分類器)，則採用 Matlab 2015(a)架設與實作，資料庫採用 MySQL，如表 3。

表 3 實驗平台環境

<table>
<tr><td>心率帶</td><td>ALATECH © - CS010</td></tr>
<tr><td>行動裝置平台</td><td>Android 4.4 (API 19)<br>JDBC Driver</td></tr>
<tr><td>HRV 分析平台</td><td rowspan="2">Matlab 2015(a)<br>JDBC Driver</td></tr>
<tr><td>運動疲勞偵測模型(包含機器學習分類器)</td></tr>
<tr><td>後端資料庫</td><td>MySQL</td></tr>
<tr><td>作業系統與硬體規格</td><td>Windows 7、Intel i5、8G ram</td></tr>
</table>

**(二)、實驗案例**

本研究主要包含四種實驗案例，前三種實驗皆是無特徵萃取的 HRV 量測指標，並採用不同的線性與非線性分析組合作為案例，最後一種則是運用本研究提出之特徵萃取方法的案例，如表 4 所示。

表 4 實驗案例

| 案例編號 | 特徵 |
|---|---|
| 實驗 1 | 無特徵萃取的 HRV 量測指標：時域分析、非線性分析(DFA、ApEn 與 SampEn)。 |
| 實驗 2 | 無特徵萃取的 HRV 量測指標：頻域分析、非線性分析(DFA、ApEn 與 SampEn)。 |
| 實驗 3 | 無特徵萃取的所有 HRV 量測指標：時域分析、頻域分析、非線性分析。 |
| 實驗 4 | 本研究方法：運動疲勞偵測模型結合特徵萃取後的 HRV 量測指標。 |

**(三)、實驗收集**

本研究實際量測五種不同生醫訊號(HR 與 RR-interval)，包含運動前、緩和運動中、劇烈運動中、緩和運動後與劇烈運動後，以每 5 分鐘為一筆量測單位。心率帶與偵測系統的取樣頻率為 0.5Hz，資料總數為 148 筆，各個運動狀態的資料數如表 5 所示。五種運動狀態執行前準備與運動行為如 3.2.2 節所述。

表 5 資料集

| 運動狀態 | 筆數 | 百分比 |
|---|---|---|
| 運動前 | 42 | 28.4% |
| 劇烈運動中 | 40 | 27% |
| 緩和運動中 | 12 | 8.1% |
| 劇烈運動後 | 42 | 28.4% |
| 緩和運動後 | 12 | 8.1% |
| 總和 | 148 | 100% |

## 二、實驗結果分析

本研究根據不同的 HRV 分析組合以及本研究方法所提出的運動疲勞偵測模型進行運動疲勞狀態辨識之實驗。其中實驗 1 至實驗 3 為沒有經過本研究方法的 HRV 分析組合，而實驗 4 則是經過本研究方法後所萃取出的 HRV 量測指標進行狀態辨識，為本研究的核心實驗結果。根據實驗結果顯示，本研究方法所提出的運動疲勞偵測模型其辨識準確率最高，平均準確率達 97.97%，實驗 1 至實驗 3 的平均準確率分別為 92.23%、88.29%與 91.89%，各分類演算法與平均準確率如表 6 所示。在原有文獻方法中，採用單一的 HRV 分析法，其 KNN-1 的辨識準確率為 66.66%，KNN-5 的辨識準確率為 66.66%，SVM 的辨識準確率為 83.33%，整體平均準確率為 72.23%。因此，本研究提出的方法較原有文獻的方法提高 25.74%的平均辨識準確率。

表 6 實驗結果準確率總表

| 分類器 / 實驗案例 | KNN-1 | KNN-5 | SVM | Neural Network | Naïve Bayes | Decision Tree | 平均 |
|---|---|---|---|---|---|---|---|
| 文獻方法 (僅時域分析) | 66.66% | 66.66% | 83.33% | - | - | - | 72.23% |
| 實驗 1 | 87.16% | 83.78% | 97.30% | 93.92% | 93.92% | 97.30% | 92.23% |
| 實驗 2 | 75.68% | 74.32% | 94.59% | 94.59% | 93.24% | 97.30% | 88.29% |
| 實驗 3 | 85.81% | 84.46% | 95.27% | 94.59% | 93.92% | 97.30% | 91.89% |
| 實驗 4 (本研究方法) | 97.97% | 97.30% | 97.97% | 98.65% | 98.65% | 97.30% | 97.97% |

此外，基於運動疲勞絕多數紀錄在劇烈運動緩合運動後，故可以由一元多次線性回歸(linear regression)方式求得趨勢線，以及線性回歸方程式，用以評估當下的運動疲勞狀態，並根據運動持續的總時間(秒)與心率值來推算未來的運動疲勞狀態。圖 6 為真實運動紀錄，主要紀錄從劇烈運動至劇烈運動後的心率表現與經過的總時間(秒)。劇烈運動中心率介於 120-181/bpm，持續時間達 70 分鐘，此部分較無運動疲勞紀錄；劇烈運動後心率介於

95-106/bpm，持續時間達 80 分鐘，此部分有最多的運動疲勞紀錄。而線性回歸方程如公式(23)所示，$x$ 為總運動時間(秒)，$y$ 為心率表現，因此可推算出 $y$ 值是介於何種運動狀態的心率表現。

$$y = f(x) = -0.000000000000012091237x^4 + 0.0000000003230596148x^3 - 0.000028960595514084x^2 + 0.0888544794070043x + 79.54454 \tag{23}$$

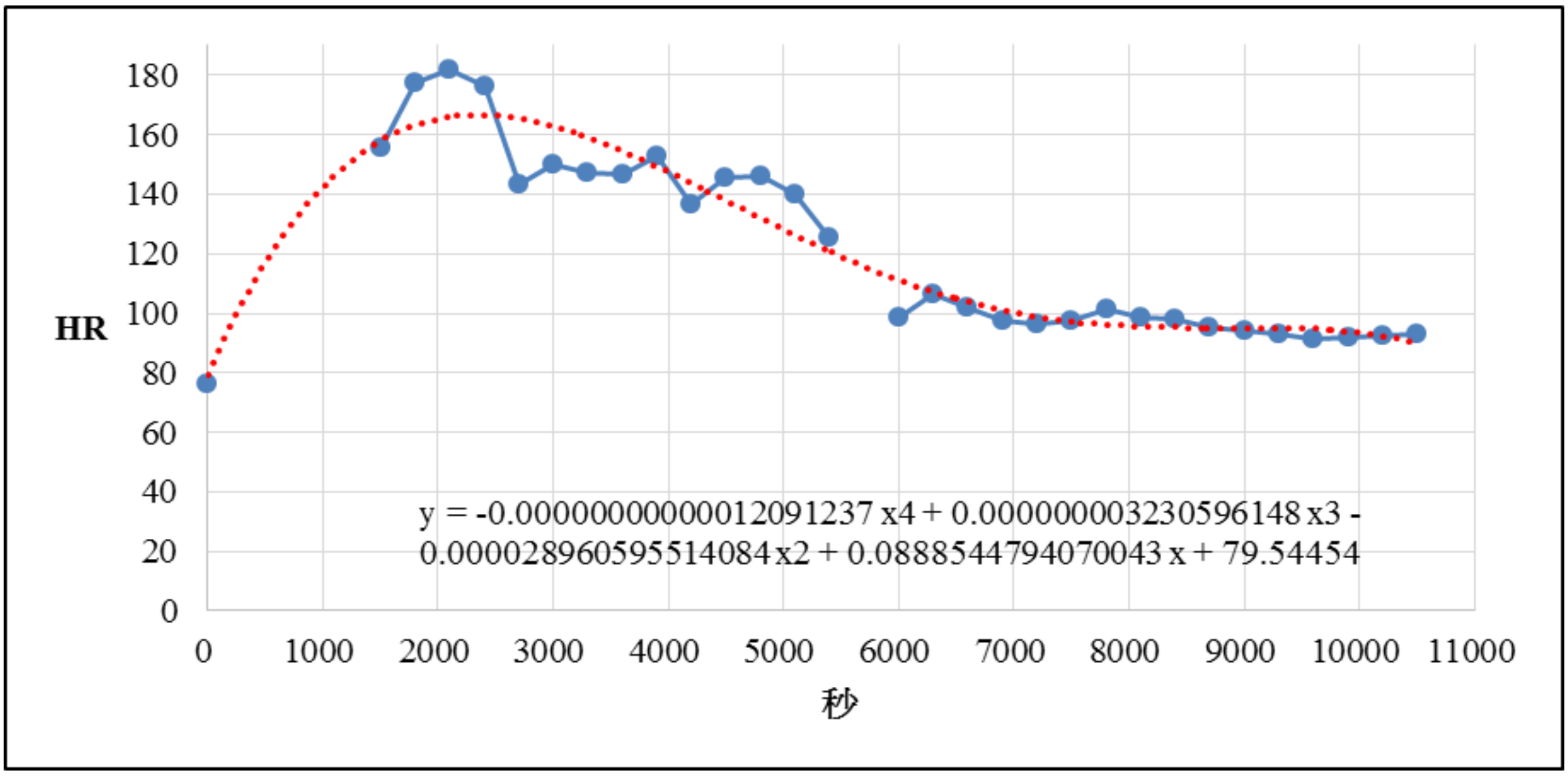

圖 6 運動疲勞趨勢線與線性回歸方程式

# 伍、結論與建議

本研究提出一個運動疲勞偵測模型，分析每個特徵因子對每個類別的重要性。透過量測兩個常態分佈特徵因子的交集面積，並由層次分析法計算所有交集面積之權重值，萃取出有價值的 HRV 量測指標作為辨別運動疲勞狀況的因子，最後，依據機器學習分類演算法判別運動疲勞狀態。根據實驗結果顯示，本研究方法所提出的運動疲勞偵測模型其辨識準確率最佳，平均準確率達 97.97%；在沒有使用本研究方法的實驗 1 至實驗 3 的平均準確率分別為 92.23%、88.29%與 91.89%；而原有文獻的方法中，採用單一的時域分析法，整體的平均準確率為 72.23%。因此，本研究提出的方法較原有文獻的方法提高 25.74%的辨識準確率，讓 MeanHR、MeanRR 與 VLF 等較高權重的 HRV 量測指標更能顯著判別個人的運動疲勞狀況。

在未來研究中，可朝向非常態分佈的資料進行分析和特徵萃取，以及可延伸運動疲勞預測模型，再依據預測資訊產生告警，以避免使用者發生運動傷害。

# 參考文獻